\documentclass[runningheads]{llncs}

\usepackage{eccv}

\usepackage{eccvabbrv}

\usepackage{graphicx}
\usepackage{booktabs}

\usepackage{graphicx}
\usepackage{booktabs}
\usepackage{multirow}
\usepackage{bm}

\usepackage[accsupp]{axessibility}  

\pdfoutput=1

\usepackage{hyperref}

\usepackage{orcidlink}

\begin{document}

\title{Frequency-Aware Guidance for Blind Image Restoration via Diffusion Models} 

\titlerunning{FAG-Diff: Frequency-Aware Guidance Diffusion}

\author{Jun Xiao\inst{1} \and
Zihang Lyu \inst{1} \and
Hao Xie\inst{1} \and
Cong Zhang\inst{1} \and
Yakun Ju\inst{2} \and
Changjian Shui\inst{3} \and
Kin-Man Lam\inst{1}}

\authorrunning{Jun Xiao et al.}

\institute{The Hong Kong Polytechnic University, Hong Kong SAR, China \and
Nanyang Technological University, Singapore \and
Vector Institute, Canada\\
\email{\{jun.xiao, zihang.lyu, carry-h.xie, cong-clarence.zhang\}@connect.polyu.hk} \\
\email{kelvin.yakun.ju@gmail.com}\quad \email{changjian.shui@vectorinstitute.ai}\quad  \email{kin.man.lam@polyu.edu.hk}}

\maketitle

\begin{abstract}
Blind image restoration remains a significant challenge in low-level vision tasks. Recently, denoising diffusion models have shown remarkable performance in image synthesis. Guided diffusion models, leveraging the potent generative priors of pre-trained models along with a differential guidance loss, have achieved promising results in blind image restoration. However, these models typically consider data consistency solely in the spatial domain, often resulting in distorted image content. In this paper, we propose a novel frequency-aware guidance loss that can be integrated into various diffusion models in a plug-and-play manner. Our proposed guidance loss, based on 2D discrete wavelet transform, simultaneously enforces content consistency in both the spatial and frequency domains. Experimental results demonstrate the effectiveness of our method in three blind restoration tasks: blind image deblurring, imaging through turbulence, and blind restoration for multiple degradations. Notably, our method achieves a significant improvement in PSNR score, with a remarkable enhancement of 3.72\,dB in image deblurring. Moreover, our method exhibits superior capability in generating images with rich details and reduced distortion, leading to the best visual quality.
  \keywords{Diffusion Models \and Posterior Sampling\and Zero-shot Restoration}
\end{abstract}

\section{Introduction}
\label{sec:intro}

Image restoration (IR) tasks, such as image denoising \cite{xiao2021bayesian,laine2019high,chen2019real}, deblurring \cite{cho2021rethinking,zhang2020deblurring,tsai2022stripformer}, super-resolution \cite{xiao2023online,xiao2019deep,wang2020deep}, etc., aim to recover high-quality images from corrupted observations, which are classic inverse problems in image processing. While IR methods have been extensively explored over the past decades, the majority of research has concentrated on non-blind restoration techniques. However, blind image restoration remains a substantial challenge and holds significant industrial value.

\begin{figure}[t]
    \centering
    \includegraphics[width=0.8\linewidth]{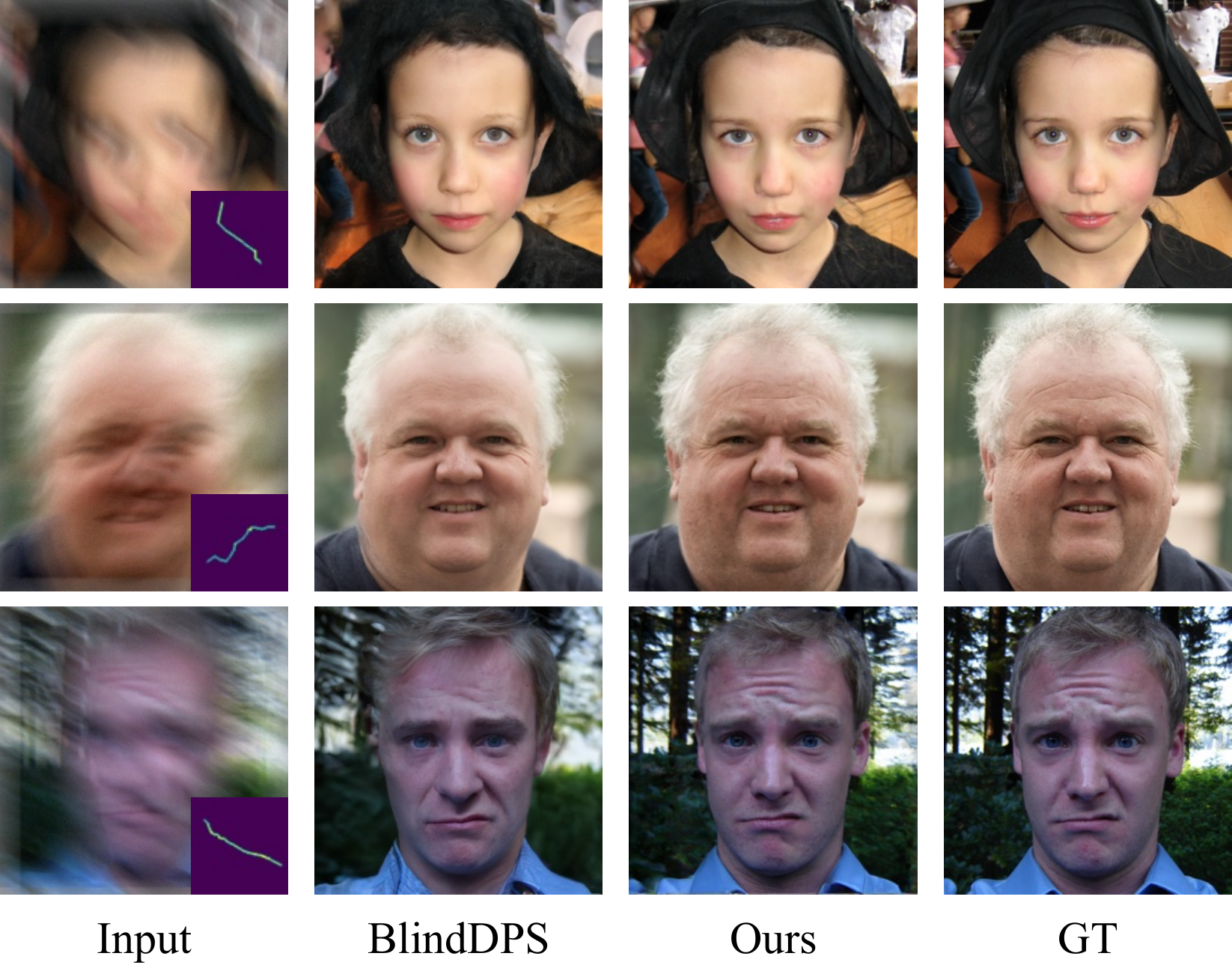}
    \caption{Visualization of the blurry images, motion kernels, images generated by BlindDPS \cite{chung2023parallel} and ours, and the corresponding ground-truth images.}
    \label{fig:intro}
\end{figure}

Recent advancements in denoising diffusion models have shown their remarkable ability in image synthesis. Inspired by the inherent generative prior provided by diffusion models, various diffusion-based IR methods have emerged. Notably, zero-shot diffusion-based IR models have gained significant attention. These models typically leverage pre-trained diffusion models and integrate differentiable loss functions into the sampling process to generate high-quality images from degraded inputs. The incorporation of these differentiable losses and the maintenance of data consistency are substantial in conditional generation, steering the sampling trajectory towards producing desirable content. MCG \cite{chung2022improving} employs a projection operator to constrain the gradient within the data manifold during the sampling process, ensuring data consistency in the reconstruction. In contrast, DPS \cite{chung2022diffusion} discards the projection operator and computes the gradient based on the estimated clean images through the linear imaging process. IIGDM \cite{song2022pseudoinverse} introduces a unified framework for image inverse problems utilizing the Moore-Penrose pseudoinverse. However, these approaches primarily address non-blind image inverse problems where the forward degradation process is known. To tackle blind image inverse problems, BlindDPS \cite{chung2023parallel} extends the DPS framework and proposes a method to estimate clean images and degradation kernels with two reverse diffusion processes. Despite these advancements, BlindDPS struggles to generate images with rich textures, often resulting in distorted image content, as depicted in Fig.~\ref{fig:intro}. Consequently, there is still significant potential for improving zero-shot diffusion-based models in blind image restoration.

In this paper, we propose a novel frequency-aware guidance loss for blind image restoration, designed to seamlessly integrate with diverse pre-trained diffusion models in a zero-shot manner. Unlike previous approaches \cite{chung2022diffusion,chung2022improving,chung2023parallel,song2022pseudoinverse}, our proposed guidance loss optimizes data consistency in both spatial and frequency domains concurrently. Building upon previous methods that consider content consistency solely in the spatial domain, our approach extends to regulate the high-frequency components of generated images, obtained through discrete wavelet transform, during the sampling process. This dual-domain optimization ensures content consistency between estimated and ground-truth images across the spatial and frequency domains. Furthermore, as high-frequency components significantly influence image perceptual quality \cite{deng2019wavelet,xiao2021balanced}, our method allows for flexible control over both reconstruction and perceptual quality by adjusting weights assigned to high-frequency guidance during sampling.

The main contributions of this paper are summarized as follows:
\begin{enumerate}
    \item We propose a frequency-aware guidance loss, based on the discrete wavelet transform, which can be incorporated into pre-trained diffusion models in a plug-and-play manner for blind image restoration.

    \item Unlike previous methods, our method simultaneously optimizes content consistency between the estimated and ground-truth images in the spatial and frequency domains, enabling flexible control over the balance between reconstruction and perceptual quality. 

    \item Experiment results show that our method effectively improves performance in three blind restoration tasks: blind image deblurring, imaging through turbulence, and blind restoration for multiple degradations. Notably, our method achieves a substantial improvement of 3.72dB in PSNR score for blind image deblurring. Furthermore, when compared with other promising methods, our method shows superior capability in generating images with rich details and reduced distortion.
\end{enumerate}

\section{Related Works}

\subsection{Image Restoration based on Diffusion Models}
As diffusion models have demonstrated remarkable efficacy in image synthesis, numerous studies tend to investigate powerful generative priors learned from pre-trained diffusion models and apply differentiable loss functions to guide the conditional generation for IR problems. These methods are known as zero-shot diffusion-based IR methods, which introduce external loss functions to generate high-quality images, without additional training. Within this realm of research, three main categories emerge: projection-based methods \cite{choi2021ilvr,kupyn2019deblurgan,chung2022come}, decomposition-based methods \cite{kawar2021snips,kawar2022denoising,kawar2022jpeg,wang2022zero}, and posterior sampling-based methods \cite{chung2022diffusion,fei2023generative,song2022pseudoinverse,fabian2023diracdiffusion,zhang2023towards,chung2022improving}.

Projection-based methods focus on performing conditional generation during the sampling process by integrating features extracted from the given degraded images. For instance, RePaint \cite{lugmayr2022repaint} considers the inpainting mask as the projection operator and applies the masked images at each sampling step. ILVR \cite{choi2021ilvr} maintains content consistency between the estimated and generated images by employing low-frequency filters in the sampling process. Unlike projection-based methods, SNIPS \cite{kawar2021snips} and DDRM \cite{kawar2021snips}, two classic decomposition-based diffusion models, utilize singular value decomposition (SVD) to conduct diffusion processes in the spectral domain, yielding promising results. Furthermore, DDNM \cite{wang2022zero} and Cheng \etal \cite{cheng2023null} introduce range-space decomposition in the sampling process, ensuring data consistency in the range space while the null space handles the reconstruction performance of the generated images. Inspired by guided diffusion models \cite{dhariwal2021diffusion}, general differential loss functions can be applied to pre-trained diffusion models in a plug-and-play manner, which serves as guidance of sampling trajectory for generating high-quality images. This is equivalent to estimating the posterior distribution of high-quality images given their corrupted observations through Bayesian posterior sampling. When approximating the underlying posterior distribution of clean images, MCG \cite{chung2022improving} and DPS \cite{chung2022diffusion} consider point estimation and estimate the gradient with task-dependent loss functions. Specifically, MCG constrains the gradient within the data manifold to ensure data consistency, while DPS computes the gradient based on the linear forward imaging process. Moreover, IIGDM \cite{song2022pseudoinverse} proposes a unified form for linear, nonlinear, and differentiable inverse problems using the Moore-Penrose pseudoinverse of the degradation function. Building on DPI \cite{sun2021deep, sun2022alpha}, Feng \etal \cite{feng2023score} define a family of distributions through RealNVP normalizing flow, optimized with minimal KL-divergence between the true posterior and the estimated distribution. All the methods mentioned above solely focus on IR problems with known degradation kernels, BlindDPS \cite{chung2023parallel} extends DPS \cite{chung2022diffusion} and proposes the utilization of two reserve diffusion processes to estimate the clean images and degradation kernels simultaneously. However, this method often generates blurry images with distorted content. 

\subsection{Frequency Learning for Image Restoration}
Deep frequency learning has demonstrated its effectiveness in image restoration and enhancement. Liu et al. \cite{liu2018multi} utilized 2D discrete wavelet transform (DWT) during the downsampling process to accelerate and improve the performance in image restoration. Similarly, Cui et al. \cite{cui2023image} addressed the discrepancy between sharp and degraded images by proposing an informative frequency selection method based on discrete wavelet transform, resulting in improved performance across various restoration tasks. Wu et al. \cite{wu2022blind} introduced a consistent network for blind image restoration, employing cycle adversarial losses to ensure data consistency in both spatial and frequency domains. Qiao et al. \cite{qiao2023learning} proposed a physically oriented generative adversarial network for unpaired image restoration, which utilizes a depth-density prior to regularizing generated content in the frequency domain through Fourier transform. Rather than applying frequency operators globally, Yoo et al. \cite{yoo2018image} suggested estimating the frequency distribution of local patches for image restoration. Similarly, Xie et al. \cite{xie2021learning} proposed a frequency-aware network to adaptively learn different frequency characteristics of local regions for accelerating image super-resolution. Taking inspiration from dynamic filtering networks, Magid et al. \cite{magid2021dynamic} introduced a dynamic high-pass filtering method to extract high-frequency features for enhancing image super-resolution. Li et al. \cite{li2023fsr} proposed a general frequency-oriented framework for accelerating the processing speed, which simultaneously learns features from both spatial and frequency domains.

In this paper, we propose a frequency-aware guidance loss based on the wavelet transform, which significantly differs from the previous works \cite{liu2018multi,cui2023image}. Our proposed loss utilizes the wavelet transform in three perspectives: 1). Our method can be seamlessly integrated with pre-trained diffusion models in a plug-and-play fashion for blind image restoration, without requiring additional training or fine-tuning. 2). While \cite{liu2018multi} utilized the wavelet transform within the pooling operator for image restoration, our approach applies frequency guidance at each diffusion step. This ensures frequency consistency in the wavelet domain for the generated images. 3). The work in \cite{cui2023image} introduced a frequency regularization term during training. In contrast, our method is training-free and operates effectively with pre-trained diffusion models at each sampling step.

\section{Methodology}

\subsection{Preliminary: Diffusion Models}

Denoising diffusion probabilistic models (DDPM) \cite{ho2020denoising,sohl2015deep} are a class of generative models that demonstrate powerful capabilities in image generation based on stochastic gradient Langevin dynamics \cite{welling2011bayesian}. These models mainly consist of forward and backward diffusion processes.

\textbf{Forward Diffusion Process}. Given the input image, denoted by $\mathbf{x}_0$, the forward process of diffusion models typically involves a Gaussian Markov Chain that iteratively corrupts the input image with additive white Gaussian noise, leading to a sequence of noisy images denoted by $\{\mathbf{x}_1, \ldots, \mathbf{x}_T\}$, where the input image gradually becomes indistinguishable from Gaussian noise over $T$ time steps. Formally, the forward diffusion process is defined as follows:
\begin{equation}
    q(\mathbf{x}_{1},\cdots, \mathbf{x}_{T}\vert \mathbf{x}_{0})=\prod_{t=1}^{T}q(\mathbf{x}_{t}\vert \mathbf{x}_{t-1}),
\end{equation}
where $q(\mathbf{x}_{t}\vert \mathbf{x}_{t-1})=\mathcal{N}(\mathbf{x}_{t}; \sqrt{1-\beta_{t}}\mathbf{x}_{t-1}, \beta_{t}\mathbf{I})$ denotes the transition kernel at the $t$-th time step, and $\beta_{t}$ is the variance schedule at the $t$-th step. Notably, the noise schedule must be delicately designed to ensure the generation of high-quality image content. As demonstrated by Ho\etal \cite{ho2020denoising}, we can efficiently compute the noisy image $\mathbf{x}_{t}$ from the input image $\mathbf{x}_{0}$ as follows:
\begin{equation}
    \mathbf{x}_{t} = \sqrt{\bar{\alpha}_{t}}\mathbf{x}_{0} + \sqrt{1-\bar{\alpha}_{t}}\mathbf{\epsilon},
\end{equation}
where $\bm{\epsilon}\sim \mathcal{N}(0, I), \alpha_{t}=1-\beta_{t}$, and $\bar{\alpha}_{t}=\prod_{i=1}^{t}\alpha_{i}$. Consequently, we obtain an alternative formulation for the transition kernel: $q(\mathbf{x}_{t}\vert \mathbf{x}_{0})=\mathcal{N}(\mathbf{x}_{t}; \sqrt{\bar{\alpha}_{t}}\mathbf{x}_{0}, (1-\bar{\alpha}_{t})\mathbf{I})$, with $\bar{\alpha}_{t}$ approaching to $0$ as the value of $T$ becomes large.

\textbf{Backward Diffusion Process}. Typically, the reverse of the forward process is intractable. Instead, DDPM learns this reverse process using a neural network, which models the reverse process as a Gaussian distribution defined as follows:
\begin{equation}
    p_{\mathbf{\theta}}(\mathbf{x}_{0},\cdots, \mathbf{x}_{T-1}\vert \mathbf{x}_{T}) = \prod_{t=1}^{T}p_{\mathbf{\theta}}(\mathbf{x}_{t-1}\vert \mathbf{x}_{t}),
\end{equation}
where $p_{\mathbf{\theta}}(\mathbf{x}_{t-1}\vert \mathbf{x}_{t})=\mathcal{N}(\mathbf{x}_{t-1}; \mathbf{\mu}_{\bm{\theta}}(\mathbf{x}_{t}, t), \Sigma_{\mathbf{\theta}}I)$. As suggested by Ho \etal \cite{ho2020denoising}, the variance $\Sigma_{\mathbf{\theta}}$ is set to constant, while the mean $\mathbf{\mu}_{\bm{\theta}}(\mathbf{x}_{t}, t)$ is learned by a neural network parameterized by $\mathbf{\theta}$, which is defined as follows:
\begin{equation}
    \mathbf{\mu}_{\mathbf{\theta}}(\mathbf{x}_{t}, t)=\frac{1}{\sqrt{\alpha_{t}}}\left(\mathbf{x}_{t}-\frac{1-\alpha_{t}}{\sqrt{1-\bar{\alpha}_{t}}}\mathbf{\epsilon}_{\theta}(\mathbf{x}_{t}, t)\right).
\end{equation}
As a result, the generative process is defined as follows:
\begin{equation}
\mathbf{x}_{t-1}  =  \frac{1}{\sqrt{\alpha_{t}}}\left(\mathbf{x}_{t}-\frac{1-\alpha_{t}}{\sqrt{1-\bar{\alpha}_{t}}}\mathbf{\epsilon}_{\theta}(\mathbf{x}_{t}, t)\right) + \mathbf{\sigma}_{t}\mathbf{z},
\end{equation}
where $\mathbf{z}\sim \mathcal{N}(0, \mathbf{I})$. Intuitively, given an input noisy sample $\mathbf{x}_{T}\sim\mathcal{N}(0, \mathbf{I})$, the backward process performs denoising on the noisy sample and iteratively transfers it back to the clean images.

\subsection{Guided Diffusion Models}
By leveraging the powerful generative prior inherent in pre-trained diffusion models, numerous research works propose to apply differentiable loss functions to pre-trained diffusion models in a plug-and-play manner for controllable generation, without requiring additional training. In zero-shot diffusion-based IR models, the objective is to learn the conditional score function $\nabla_{\mathbf{x}_{t}}\log p_{t}(\mathbf{x}_{t}|\mathbf{y})$, where $\mathbf{y}$ represents the degraded image. By Bayes' rule, it can be decomposed as follows:
\begin{equation}
    \nabla_{\mathbf{x}_{t}}\log p_{t}(\mathbf{x}_{t}\vert \mathbf{y}) =\nabla_{\mathbf{x}_{t}}\log p_{t}(\mathbf{x}_{t}) + \nabla_{\mathbf{x}_{t}}\log p_{t}(\mathbf{y}\vert \mathbf{x}_{t}),
\end{equation}
where the first term, $\nabla_{\mathbf{x}_{t}}\log p_{t}(\mathbf{x}_{t})$, represents the unconditional score function computed by the unconditional diffusion model. The second term, $\nabla{\mathbf{x}_{t}}\log p_{t}(\mathbf{y}|\mathbf{x}_{t})$, corresponds to the adversarial gradient calculated by an external pre-trained classifier based on the intermediate output $\mathbf{x}_{t}$. However, training such classifiers for noisy images at each sampling step is impractical and unsuitable for image restoration problems. To address this issue, DPS \cite{chung2022diffusion} proposes to approximate the posterior distribution with the estimated clean image, computed as follows:
\begin{equation}
    \nabla_{\mathbf{x}_{t}} \log p_{t}(\mathbf{y}\vert \mathbf{x}_{t}) \approx \nabla_{\mathbf{x}_{t}}\log p_{t}(\mathbf{y}\vert \hat{\mathbf{x}}_{0})=\nabla_{\mathbf{x}_{t}} \left\Arrowvert \mathbf{y} - \mathbf{k}\ast \mathbf{\hat{x}}_{0}\right\Arrowvert_{2}^{2},
\end{equation}
where $\mathbf{k}$ represents the degradation kernel, and $\mathbf{\hat{x}}_{0}=\frac{1}{\sqrt{\bar{\alpha}_{t}}}\left(\mathbf{x}_{t}-(1-\bar{\alpha}_{t})\mathbf{\epsilon}_{\theta}(\mathbf{x}_{t}, t)\right)$ represents the estimated clean images derived from Tweedie's formula. Instead of utilizing an external classifier, it computes the adversarial gradient based on the linear image forward process. To handle blind image restoration, BlindDPS \cite{chung2023parallel} extends this approach and computes the adversarial gradient jointly conditioned on the estimated images and degradation kernels, which is defined as follows:
\begin{equation}
    \nabla_{\mathbf{x}_{t}}\log p_{t}(\mathbf{y}\vert \mathbf{\hat{x}}_{0}, \mathbf{\hat{k}}_{0})=\nabla_{\mathbf{x}_{t}} \left\Arrowvert \mathbf{y} - \mathbf{\hat{k}}_{0}\ast \mathbf{\hat{x}}_{0}\right\Arrowvert_{2}^{2},
\end{equation}
where $\ast$ denotes the convolutional operator, and $\mathbf{\hat{k}}_{0}$ represents the estimated degradation kernel obtained through another reverse diffusion process. However, BlindDPS often produces distorted image content, leading to deteriorated performance. Therefore, there is significant potential for improvement.

\subsection{Posterior Sampling with Frequency Guidance}
In this paper, we propose a frequency-aware guidance loss that simultaneously optimizes the data consistency of the estimated images in both the spatial and frequency domains. Similar to previous studies \cite{chung2022diffusion,chung2022improving,song2022pseudoinverse}, our proposed method can be seamlessly incorporated into pre-trained diffusion models in a plug-and-play manner, guiding the sampling trajectory with frequency information.

In BlindDPS \cite{chung2023parallel}, the estimated degraded observation $\mathbf{\hat{y}}=\mathbf{\hat{k}}_{0}\ast \mathbf{\hat{x}}_{0}$ is computed based on the estimated clean image $\hat{\mathbf{x}}_{0}$ and the estimated degradation kernel $\mathbf{\hat{k}}_{0}$. In our method, before computing the data consistency in the spatial domain, we first apply 2D DWT with four convolutional filters, i.e., one low-frequency filter $k_{\text{LL}}$ and three high-frequency filters, denoted by $k_{\text{LH}}$, $k_{\text{HL}}$, and $k_{\text{HH}}$, to decompose the estimated degraded observation $\mathbf{\hat{y}}$ into four frequency subbands, which are denoted as $\mathbf{\hat{y}}_{\text{LL}}$, $\mathbf{\hat{y}}_{\text{Lh}}$, $\mathbf{\hat{y}}_{\text{HL}}$, and $\mathbf{\hat{y}}_{\text{HH}}$, respectively. In our method, we utilize Haar wavelet, and the four wavelet filters are defined as follows:
\begin{equation}
\begin{split}
    k_{\text{LL}} = \begin{bmatrix}
        1 & 1 \\
        1 & 1
    \end{bmatrix},\quad
    k_{\text{LH}} = \begin{bmatrix}
        -1 & -1 \\
        1 & 1
    \end{bmatrix},\quad
    k_{\text{HL}} = \begin{bmatrix}
        -1 & 1 \\
        -1 & 1
    \end{bmatrix},\quad
    k_{\text{HH}} = \begin{bmatrix}
        1 & -1 \\
        -1 & 1
    \end{bmatrix}. 
\end{split}
\end{equation}
It is evident that $k_{\text{LL}}$, $k_{\text{LH}}$, $k_{\text{HL}}$, and $k_{\text{HH}}$ are orthogonal to each other. The operation of DWT is defined as $\mathbf{\hat{y}}_{\text{LL}}=(k_{\text{LL}}\ast \mathbf{\hat{y}})\downarrow_{2}$, $\mathbf{\hat{y}}_{\text{LH}}=(k_{\text{LH}}\ast \mathbf{\hat{y}})\downarrow_{2}$, $\mathbf{\hat{y}}_{\text{HL}}=(k_{\text{HL}}\ast \mathbf{\hat{y}})\downarrow_{2}$, and $\mathbf{\hat{y}}_{\text{HH}}=(k_{\text{HH}}\ast \mathbf{\hat{y}})\downarrow_{2}$, where $\downarrow_{2}$ represents the downsampling operator with the scaling factor of $2$. In the implementation, DWT is equivalent to applying four convolutional filters with a stride of $2$. Specifically, the pixel values at the position $(i, j)$ of $\mathbf{\hat{y}}_{\text{LL}}, \mathbf{\hat{y}}_{\text{LH}}, \mathbf{\hat{y}}_{\text{HL}}$, and $\mathbf{\hat{y}}_{\text{HH}}$ are computed as follows:
\begin{align}
    \mathbf{\hat{y}}_{\text{LL}}(i, j) &= \mathbf{\hat{y}}(2i-1, 2j-1) + \mathbf{\hat{y}}(2i-1,2j) +\mathbf{\hat{y}}(2i, 2j-1) + \mathbf{\hat{y}}(2i,2j),\\
    \mathbf{\hat{y}}_{\text{LH}}(i, j) &= -\mathbf{\hat{y}}(2i-1, 2j-1) - \mathbf{\hat{y}}(2i-1,2j)+\mathbf{\hat{y}}(2i, 2j-1) + \mathbf{\hat{y}}(2i,2j), \\
    \mathbf{\hat{y}}_{\text{HL}}(i, j) &= -\mathbf{\hat{y}}(2i-1, 2j-1) + \mathbf{\hat{y}}(2i-1,2j) -\mathbf{\hat{y}}(2i, 2j-1) + \mathbf{\hat{y}}(2i,2j),\\
    \mathbf{\hat{y}}_{\text{HH}}(i, j) &= \mathbf{\hat{y}}(2i-1, 2j-1) - \mathbf{\hat{y}}(2i-1,2j) -\mathbf{\hat{y}}(2i, 2j-1) + \mathbf{\hat{y}}(2i,2j).     
\end{align}

After wavelet decomposition, the frequency-aware guidance loss $\mathcal{L}_{\text{freq}}$ is computed as follows:
\begin{equation}
    \mathcal{L}_{\text{freq}}=\Arrowvert \mathbf{y} - \mathbf{\hat{y}}\Arrowvert_{2}^{2} + \sum_{i\in \{\text{LH, HL, HH}\}}\lambda_{i}\cdot\Arrowvert \bm{y}_{i}-\hat{\bm{y}}_{i}\Arrowvert_{2}^{2},
\end{equation}
where $\lambda_{i}$ is a hyper-parameter to control the regularization strength of the high-frequency content in the generated images. A large value of $\lambda_{i}$ means that the model pays more attention to generating high-frequency details. Compared with previous studies, our proposed method enforces data consistency of the generated images in both the spatial and frequency domains. In addition, previous studies \cite{xiao2021balanced,deng2019wavelet} reveal that the main content or energy of an image concentrates on its low-frequency components, so the content quality of an image is highly related to its low-frequency information. In contrast, its high-frequency components typically involve textural information, significantly affecting the visual quality of an image. Therefore, our method provides an effective way to balance the reconstruction and perceptual quality of generated images by adjusting the weights assigned to the high-frequency guidance.

\section{Experiments and Analysis}

\subsection{Experiment on Blind Image Deblurring}
\textbf{Dataset information}. In this experiment, we utilize the FFHQ validation dataset for evaluation, following the configurations used in the previous study \cite{chung2023parallel}. Specifically, we select 10,000 images from the FFHQ dataset, and each image is pre-processed and resized to a size of $256\times 256$. 

\noindent\textbf{Implementation details}. It is worth noticing that our proposed method is training-free and can be incorporated with the pre-trained diffusion models in a plug-and-play manner. Therefore, in our implementation, we apply our method to the pre-trained diffusion models provided by Chung et al. \cite{chung2023parallel}. This model is pre-trained on a large-scale dataset and utilizes a small U-Net structure for degradation kernel prediction. In addition, the intensity values of degradation kernels are set to $0.5$ and $3.0$, respectively, for randomly generating motion and Gaussian kernels applied to each image, which are the default settings suggested by the previous research work \cite{chung2023parallel}. We empirically set the hyperparameter $\lambda$ to $0.1$ in the experiment. We compare our method with SelfDeblur \cite{ren2020neural}, MPRNet \cite{zamir2021multi}, DeblurGANv2 \cite{kupyn2019deblurgan}, PAN-DCP \cite{pan2017deblurring}, PAN-$\ell_{0}$ \cite{pan2016l_0}, and BlindDPS \cite{chung2022improving}. Among the compared methods, MPRNet and DeblurGANv2 are supervised learning methods specifically trained on large-scale datasets for image deblurring. BlindDPS is also a training-free method based on the diffusion model. To ensure a fair comparison, we utilize their open-source codes to implement these methods.

\noindent\textbf{Evaluation metrics}. To evaluate the performance of different methods, we use the peak signal-to-noise ratio (PSNR) to measure the reconstruction quality of the generated images, while the Frechet inception distance (FID) and the Learned Perceptual Image Patch Similarity (LPIPS) \cite{zhang2018unreasonable} are employed to assess the visual quality of generated images.

\begin{table}[!ht]
\centering
\tabcolsep=0.13cm
\renewcommand\arraystretch{1.2}
\caption{The average FID, LPIPS, and PSNR scores of different blind deblurring methods on the FFHQ datasets under the degradations of motion and Gaussian blur kernels. The best results are highlighted in bold.}
\begin{tabular}{|c|cccccc|}
\hline
\multirow{3}{*}{Method} & \multicolumn{6}{c|}{FFHQ ($256\times 256$)}    \\ \cline{2-7} 
                        & \multicolumn{3}{c|}{Motion}                                                          & \multicolumn{3}{c|}{Gaussian}      \\ \cline{2-7} 
                        & \multicolumn{1}{c}{FID$\downarrow$}   & \multicolumn{1}{c}{LPIPS$\downarrow$} & \multicolumn{1}{c|}{PSNR$\uparrow$}  & \multicolumn{1}{c}{FID$\downarrow$}   & \multicolumn{1}{c}{LPIPS$\downarrow$} & PSNR$\uparrow$  \\ \hline
SelfDeblur \cite{ren2020neural}             & \multicolumn{1}{c}{270.0} & \multicolumn{1}{c}{0.717} & \multicolumn{1}{c|}{10.83} & \multicolumn{1}{c}{235.4} & \multicolumn{1}{c}{0.686} & 11.36  \\ 
MPRNet \cite{zamir2021multi}                  & \multicolumn{1}{c}{111.6} & \multicolumn{1}{c}{0.434} & \multicolumn{1}{c|}{17.40} & \multicolumn{1}{c}{95.12} & \multicolumn{1}{c}{0.337} & 20.75 \\ 
DeblurGANv2 \cite{kupyn2019deblurgan}            & \multicolumn{1}{c}{220.7} & \multicolumn{1}{c}{0.571} & \multicolumn{1}{c|}{17.75} & \multicolumn{1}{c}{185.5} & \multicolumn{1}{c}{0.529} & 19.69 \\ 
PAN-DCP \cite{pan2017deblurring}                & \multicolumn{1}{c}{214.9} & \multicolumn{1}{c}{0.520} & \multicolumn{1}{c|}{15.41} & \multicolumn{1}{c}{92.70} & \multicolumn{1}{c}{0.393} & 20.50 \\ 
PAN-$\ell_{0}$ \cite{pan2016l_0}          & \multicolumn{1}{c}{242.6} & \multicolumn{1}{c}{0.542} & \multicolumn{1}{c|}{15.53} & \multicolumn{1}{c}{109.1} & \multicolumn{1}{c}{0.415} & 19.94 \\ 
BlindDPS \cite{chung2023parallel}               & \multicolumn{1}{c}{29.49} & \multicolumn{1}{c}{0.281} & \multicolumn{1}{c|}{22.24} & \multicolumn{1}{c}{27.36} & \multicolumn{1}{c}{0.233} & 24.77 \\ 
Ours               & \multicolumn{1}{c}{\textbf{25.15}} & \multicolumn{1}{c}{\textbf{0.224}} & \multicolumn{1}{c|}{\textbf{25.96}} & \multicolumn{1}{c}{\textbf{26.96}} & \multicolumn{1}{c}{\textbf{0.212}} & \textbf{26.23} \\ \hline
\end{tabular}
\label{table_1}
\end{table}

\begin{figure}[!ht]
    \centering
    \includegraphics[width=0.95\linewidth]{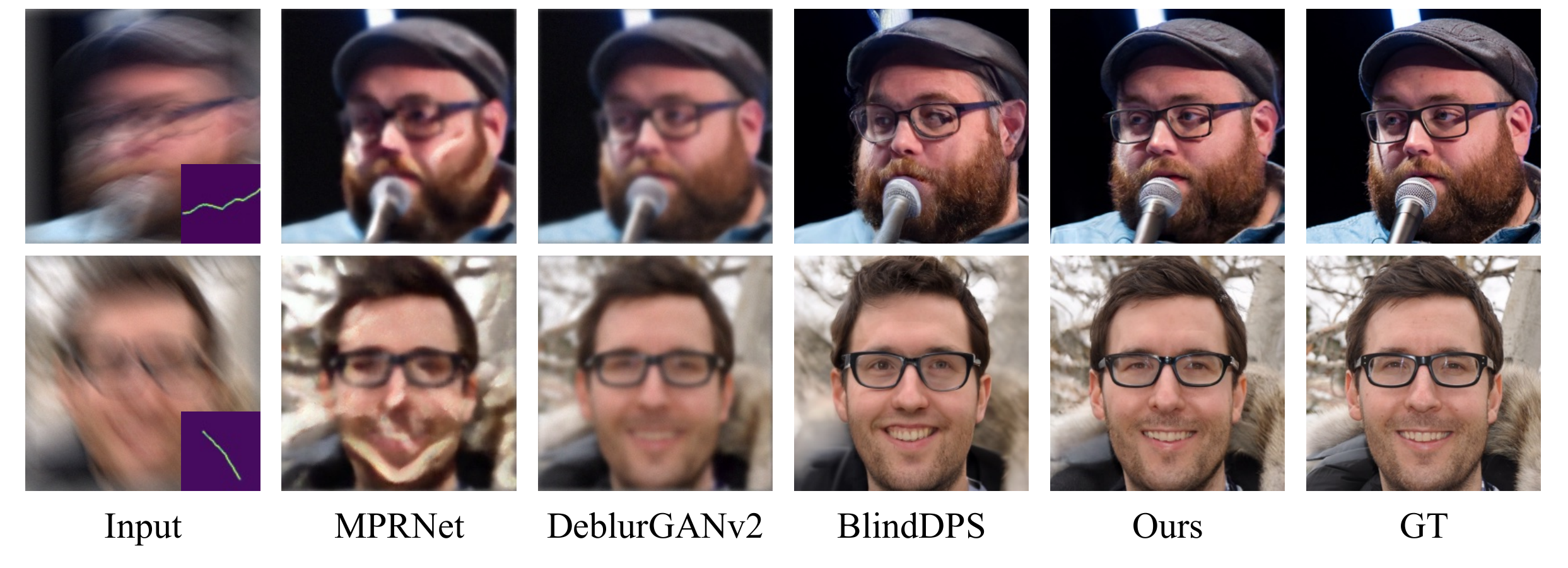}
    \caption{Visual results of the images generated by MPRNet \cite{zamir2021multi}, DeblurGANv2 \cite{kupyn2019deblurgan}, BlindDPS \cite{chung2023parallel}, and our method.}
    \label{visual_1}
\end{figure}

\noindent\textbf{Result analysis}. The average FID, LPIPS, and PSNR scores of different blind restoration methods on the FFHQ dataset with motion and Gaussian blur degradations are shown in Table \ref{table_1}. Our method achieves the best performance in various evaluation metrics on the FFHQ datasets. In particular, significant improvements are obtained in the PSNR score, with an increase of 3.72dB and 1.46dB for the motion blur kernel and Gaussian blur kernel, respectively. These results show that our method has superior generalization capability when testing images that experience domain shifts in terms of blur kernels. Furthermore, we provide visual comparisons of images generated by different methods in Fig.~\ref{visual_1}. MPRNet and DeblurGANv2 struggle to produce clean image content and exhibit undesirable artifacts. While BlindDPS can generate images with clean content, they lack consistency with ground-truth images and detailed information. In contrast, images generated by our method show reduced distortion, rich texture, and detailed information, resulting in better visual quality. This improvement can be attributed to our method which simultaneously considers the data consistency in both the spatial and frequency domains, leading to a more accurate sampling trajectory.

\subsection{Experiments on Imaging Through Turbulence}


\noindent\textbf{Dataset information}. In addition to 10,000 images from the FFHQ dataset, we include the ImageNet-1K validation dataset, which contains 50,000 images for evaluation. All testing images undergo the same pre-processing steps and are resized to dimensions of $256\times 256$. 

\noindent\textbf{Implementation details}. Similar to the previous experiment of image deblurring, we employ the same pre-trained diffusion model with the small U-Net structure for restoration and kernel prediction. Additionally, we followed the default setting by \cite{chung2023parallel} and utilized another diffusion model to predict the tile map. This diffusion model has been trained on 50,000 randomly generated tile maps, and other model configurations are kept unchanged to ensure a fair comparison. To generate the low-quality images during testing, the point spread function (PSF) is assumed to follow the Gaussian distribution. Specifically, we set the standard deviations to $4.0$ for the FFHQ dataset and $2.0$ for the ImageNet-1K validation dataset. In the experiment, we compare our method with several promising restoration models, including TSR-WGAN \cite{jin2021neutralizing}, ILVR \cite{choi2021ilvr}, MPRNet \cite{zamir2021multi}, DeblurGANv2 \cite{kupyn2019deblurgan}, and BlindDPS \cite{chung2022improving}. We utilized their publicly available source codes with default settings for evaluation. To access the reconstruction and perceptual qualities, we employed PSNR, FID, and LPIPS metrics.

\begin{table}[!ht]
\centering
\tabcolsep=0.25cm
\renewcommand\arraystretch{1.2}
\caption{The average PSNR, LPIPS, and FID scores of different image restoration models on the FFHQ dataset and the ImageNet-1K dataset. Each image in this dataset is resized to the dimension of $256\times 256$. The best results are highlighted in bold.}
\begin{tabular}{|c|ccc|ccc|}
\hline
\multirow{2}{*}{Methods} & \multicolumn{3}{c|}{FFHQ (256x256)}                             & \multicolumn{3}{c|}{ImageNet (256x256)}                         \\ \cline{2-7} 
                         & \multicolumn{1}{c}{PSNR$\uparrow$}  & \multicolumn{1}{c}{LPIPS$\downarrow$} & FID$\downarrow$   & \multicolumn{1}{c}{PSNR$\uparrow$}  & \multicolumn{1}{c}{LPIPS$\downarrow$} & FID$\downarrow$   \\ \hline
TSR-WGAN \cite{jin2021neutralizing}                & \multicolumn{1}{c}{26.29} & \multicolumn{1}{c}{0.258} & 58.30 & \multicolumn{1}{c}{17.67} & \multicolumn{1}{c}{0.369} & 69.80 \\
ILVR \cite{choi2021ilvr}                    & \multicolumn{1}{c}{21.48} & \multicolumn{1}{c}{0.370} & 65.50 & \multicolumn{1}{c}{18.09} & \multicolumn{1}{c}{0.494} & 85.21 \\ 
MPRNet \cite{zamir2021multi}                  & \multicolumn{1}{c}{19.68} & \multicolumn{1}{c}{0.411} & 116.2 & \multicolumn{1}{c}{20.34} & \multicolumn{1}{c}{0.421} & 78.24 \\
DeblurGANv2 \cite{kupyn2019deblurgan}             & \multicolumn{1}{c}{18.40} & \multicolumn{1}{c}{0.561} & 225.9 & \multicolumn{1}{c}{21.56} & \multicolumn{1}{c}{0.393} & 60.31 \\
BlindDPS  \cite{chung2022improving}               & \multicolumn{1}{c}{24.49} & \multicolumn{1}{c}{0.247} & 27.35 & \multicolumn{1}{c}{19.59} & \multicolumn{1}{c}{\textbf{0.341}} & \textbf{51.25} \\
Ours                  & \multicolumn{1}{c}{\textbf{26.37}} & \multicolumn{1}{c}{\textbf{0.214}} & \textbf{25.51} & \multicolumn{1}{c}{\textbf{21.75}} & \multicolumn{1}{c}{0.389} & 80.78 \\ \hline
\end{tabular}
\label{table_2}
\end{table}

\begin{figure*}[!ht]
    \centering
    \includegraphics[width=0.95\linewidth]{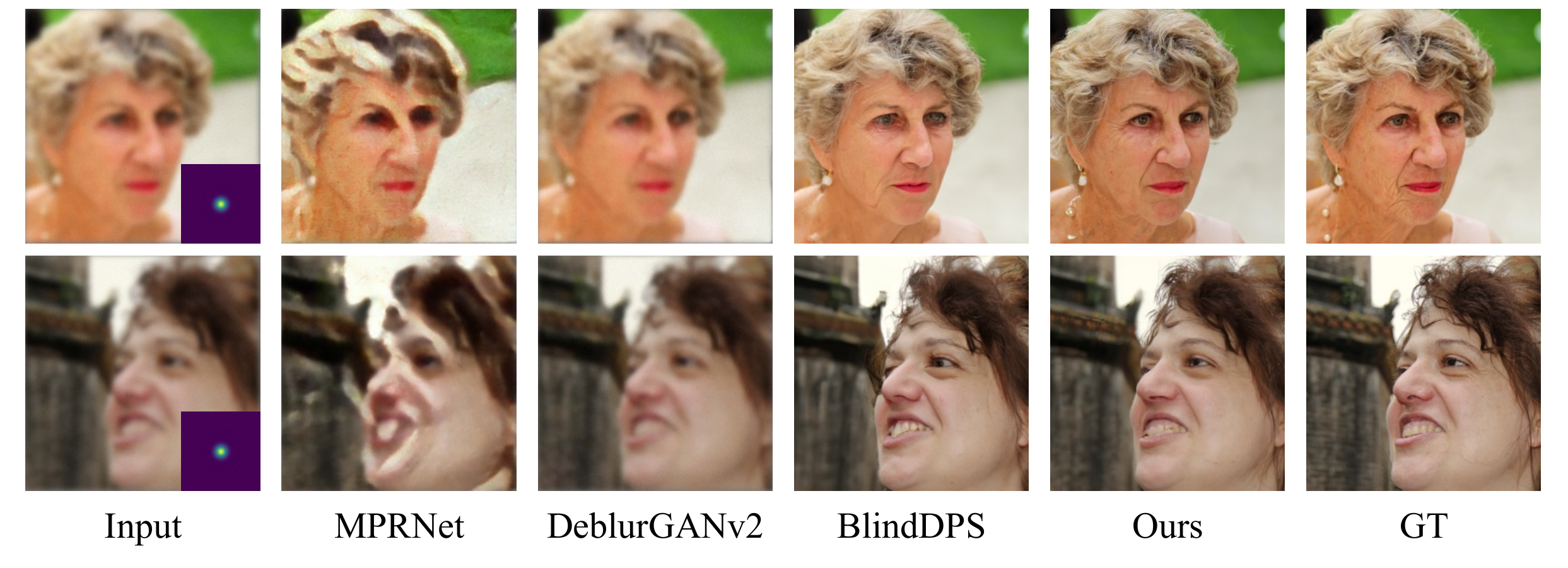}
    \caption{Visual results of the images generated by  MPRNet \cite{zamir2021multi}, DeblurGANv2 \cite{kupyn2019deblurgan}, BlindDPS \cite{chung2023parallel}, and our method.}
    \label{fig_3}
\end{figure*}

\noindent\textbf{Result Analysis}. Table~\ref{table_2} illustrates the average PSNR, FID, and LPIPS scores of different methods on the FFHQ and ImageNet-1K datasets for imaging through turbulence. The results show that our proposed method achieve significant improvements in PSNR on both datasets, with gains of 1.88dB on the FFHQ dataset and 2.19dB on the ImageNet-1K dataset, demonstrating superior reconstruction performance compared to other methods. In terms of the visual quality, our method outperforms the comparison methods in FID and LPIPS on the FFHQ dataset and achieves comparable LPIPS results on the ImageNet-1K dataset. Unlike BlindDPS, the diffusion model used in our method is pre-trained on the FFHQ dataset, resulting in a domain gap when applied to the ImageNet-1K dataset. Moreover, as shown in Fig.~\ref{fig_3}, our method exhibits a better capability in generating texture and detailed content in local areas, such as facial wrinkles.

\subsection{Experiments on Multiple Degradations}
Real-world images often suffer from various types of degradation, so we further evaluate our proposed method on images with multiple degradations, which is a more challenging setting compared to the previous experiments. Specifically, we evaluated our method on 10,000 images from the FFHQ dataset, using the same testing configurations as in the previous experiments, except for the degradation kernel. In the experiment, the input images were corrupted by motion blur and Gaussian noise (with the noise level set to 10).


\begin{table}[!ht]
\centering
\renewcommand\arraystretch{1.2}
\caption{The average PSNR, LPIPS, and FID scores of different models on the FFHQ dataset. Each image is corrupted with multiple degradations (i.e., motion blur and noise level of 10). The best results are highlighted in bold.}
\setlength{\tabcolsep}{3mm}{
\begin{tabular}{|ccccc|}
\hline
 & MPRNet \cite{zamir2021multi} & DeblurGANv2 \cite{kupyn2019deblurgan} & BlindDPS  \cite{chung2022improving} & Ours \\ \hline
PSNR$\uparrow$ & 16.19 & 17.82 & 22.19 &  \textbf{25.04}\\ 
LPIPS$\downarrow$ & 0.685 & 0.615 & 0.314 & \textbf{0.259} \\ 
FID$\downarrow$ & 266.56 & 156.35 & 32.15 &  \textbf{30.16}\\ \hline
\end{tabular}}
\label{table_degradations}
\end{table}
\begin{figure}[!ht]
    \centering
    \includegraphics[width=0.95\linewidth]{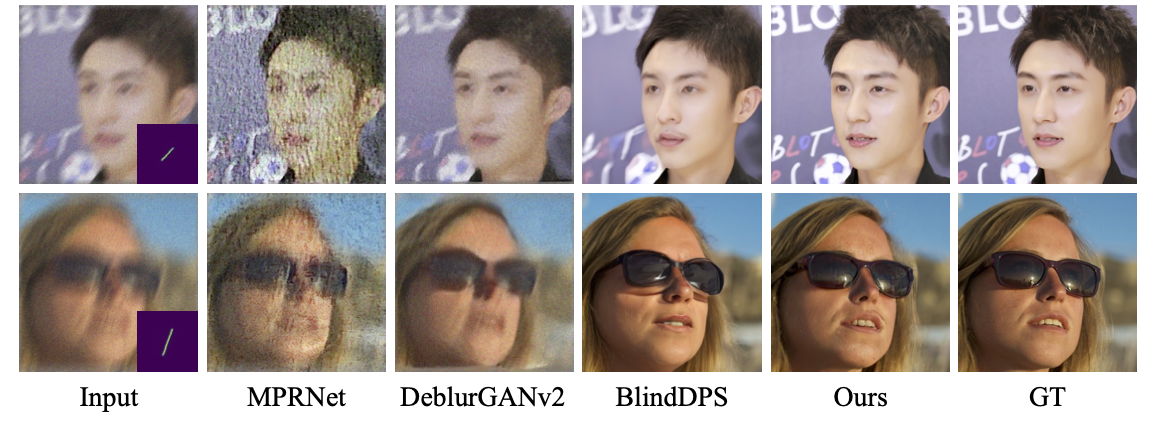}
    \caption{Visual results of the images generated by MPRNet \cite{zamir2021multi}, DeblurGANv2 \cite{kupyn2019deblurgan}, BlindDPS \cite{chung2023parallel}, and our method.}
    \label{visual_degradations}
\end{figure}
The average PSNR, LPIPS, and FID scores of different restoration models on the FFHQ dataset with multiple degradations are illustrated in Table \ref{table_degradations}. Our method significantly outperforms the compared methods across all evaluation metrics. These results demonstrate that our method effectively handles both blurry and noisy content simultaneously and is robust to a mixture of multiple degradations, which is crucial for real-world image restoration. Additionally, Fig. \ref{visual_degradations} demonstrates the images generated by different models for visual comparison. As observed, our method better preserves the identity of the subjects, resulting in images with less distortion and higher visual quality.


\subsection{Ablation Studies}
\textbf{Experiments on the hyperparameter $\lambda$}.
As our method considers data consistency in the spatial and frequency domains, achieving a balance in guided information from these two domains becomes crucial. In this experiment, we explore the impact of the hyperparameter $\lambda$, which controls the strength of high-frequency guidance loss. Specifically, we vary $\lambda$ across values of $0.01, 0.1, 0.5, 1.0$, and $5.0$, and select 100 images from the FFHQ dataset for evaluation.
\begin{table}[!ht]
\renewcommand\arraystretch{1.2}
\centering
\caption{The average PSNR, LPIPS, and FID scores of our method with different values of $\lambda$ on the selected 100 images from the FFHQ dataset. The best results are highlighted in bold.}
\setlength{\tabcolsep}{3mm}{
\begin{tabular}{|c|ccccc|}
\hline
\multirow{2}{*}{Metrics} & \multicolumn{5}{c|}{The value of $\lambda$}                                                                                                     \\ \cline{2-6} 
                         & \multicolumn{1}{c}{0.01}  & \multicolumn{1}{c}{0.1}   & \multicolumn{1}{c}{0.5}   & \multicolumn{1}{c}{1.0}   & 5.0   \\ \hline
PSNR $\uparrow$                     & \multicolumn{1}{c}{25.99} & \multicolumn{1}{c}{\textbf{26.41}} & \multicolumn{1}{c}{26.14} & \multicolumn{1}{c}{26.28} & 25.08 \\ 
LPIPS $\downarrow$                   & \multicolumn{1}{c}{0.224} & \multicolumn{1}{c}{\textbf{0.215}} & \multicolumn{1}{c}{\textbf{0.215}} & \multicolumn{1}{c}{0.222} & 0.297 \\ 
FID  $\downarrow$                    & \multicolumn{1}{c}{62.72} & \multicolumn{1}{c}{58.52} & \multicolumn{1}{c}{\textbf{56.37}} & \multicolumn{1}{c}{61.53} & 81.88 \\ \hline
\end{tabular}}
\label{table_3}
\end{table}

\begin{figure}[!ht]
    \centering
    \includegraphics[width=0.95\linewidth]{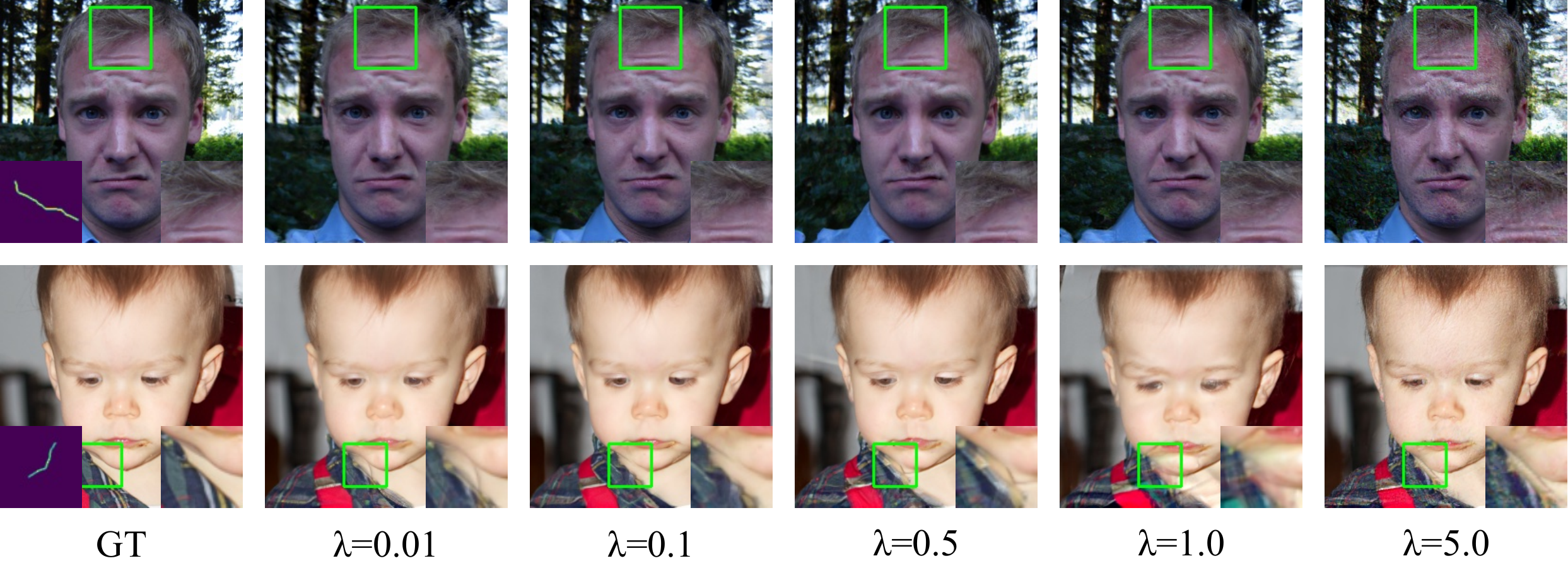}
    \caption{Visual results of our method with various values of the hyper-parameter $\lambda$, where $\lambda$ is set to $0.01, 0.1, 0.5, 1.0$, and $5.0$}
    \label{visual_3}
\end{figure}

The average PSNR, LPIPS, and FID scores of our method with different values of $\lambda$ are shown in Table \ref{table_3}. It is evident that the hyperparameter $\lambda$ significantly affects the reconstruction performance and visual quality. Specifically, when the value of $\lambda$ is set to $0.1$, $0.5$, or $1.0$, our method yields similar results, especially in FID and LPIPS scores. However, when the value of $\lambda$ is set to $0.01$ or $5.0$, the reconstruction performance of our method is seriously deteriorated. Visual comparisons of images generated by models with different $\lambda$ values are illustrated in Fig.~\ref{visual_3}. As observed, employing large values of $\lambda$ (e.g., $0.5$, $1.0$, and $5.0$) tends to produce distorted image content due to the dominance of high-frequency guidance loss. Conversely, when the value of $\lambda$ is too small (e.g., $0.01$), the model neglects local details, resulting in overly smooth regions.

\noindent\textbf{Experiments on Frequency Components}.
Our method based on the 2D DWT decomposes images into one low-frequency component, denoted by LL, and three high-frequency components, denoted by LH, HL, and HH, respectively. In this experiment, we thoroughly analyze the impact of these high-frequency components on image generation using 100 images from the FFHQ dataset. Specifically, we assess model performance by employing different combinations of frequency components, which include the spatial consistency alone denoted by $L_{2}$, $L_{2}$ + $\mathcal{H}_{1}$ components, $L_{2}$ + $\mathcal{H}_{2}$ components, and $L_{2}$ + $\mathcal{H}_{3}$ components, where $\mathcal{H}_{1} = \{\text{LH}\}, \mathcal{H}_{2} = \{\text{LH}, \text{HL}\}$, and $\mathcal{H}_{3}=\{\text{LH}, \text{HL}, \text{HH}\}$.

\begin{figure}[!ht]
    \centering
    \includegraphics[width=0.95\linewidth]{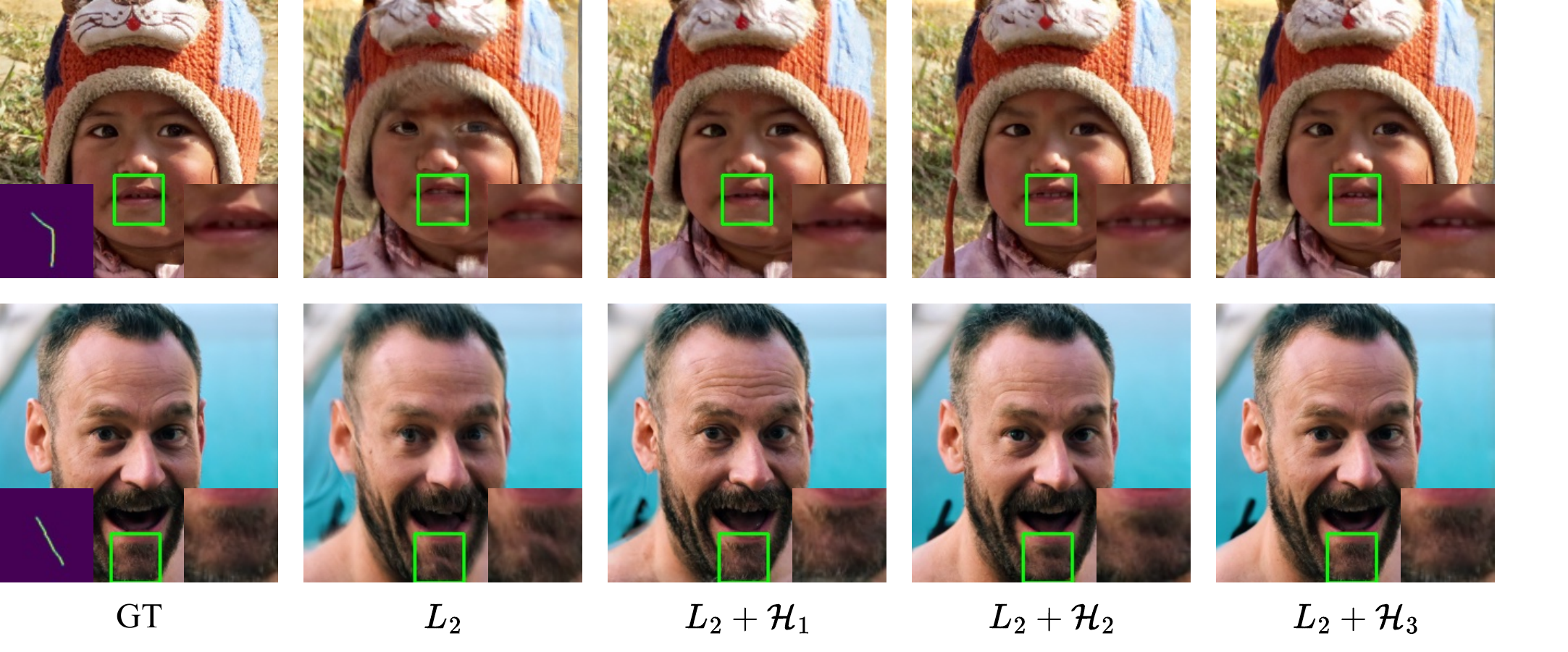}
    \caption{Visual results of the models with different combinations of frequency components, where $\mathcal{H}_{1} = \{\text{LH}\}, \mathcal{H}_{2} = \{\text{LH}, \text{HL}\}$, and $\mathcal{H}_{3}=\{\text{LH}, \text{HL}, \text{HH}\}$.}
    \label{visual_4}
\end{figure}

The average PSNR, LPIPS, and FID scores of the model with employing different frequency components are shown in Table~\ref{table_4}. From these results, we find that the reconstruction and perceptual performances are improved when incorporating more high-frequency components in the high-frequency guidance loss. The corresponding visual results shown in Fig.~\ref{visual_4} also demonstrate that images generated by models using more high-frequency components exhibit cleaner content with reduced distortion and richer local details. These findings reveal that our method has a superior capability in preserving data consistency, resulting in generated images that closely approach the ground-truth images.

\begin{table}[!ht]
\renewcommand\arraystretch{1.2}
\centering
\caption{The average PSNR, LPIPS, and FID of our methods employing different combinations of frequency components. The best results are highlighted in bold.}
\setlength{\tabcolsep}{3mm}{
\begin{tabular}{|c|c|c|c|c|}
\hline
      & $L_{2}$ & $L_{2}$+$\mathcal{H}_{1}$ & $L_{2}$+$\mathcal{H}_{2}$ & $L_{2}$+$\mathcal{H}_{3}$ \\ \hline
PSNR $\uparrow$  & 26.09   & 26.20    & 26.17       &  \textbf{26.64}         \\ 
LPIPS $\downarrow$ & 0.225   & 0.224    & 0.223       &  \textbf{0.206}         \\ 
FID $\downarrow$  & 62.42   & 59.11    & 61.14       &  \textbf{54.12}         \\ \hline
\end{tabular}}
\label{table_4}
\end{table}

\section{Conclusion}
In this paper, we propose a frequency-aware guidance loss for blind image restoration, which integrates with various pre-trained diffusion models without the need for additional training. Our proposed guidance loss concurrently optimizes data consistency in both the spatial and frequency domains. Specifically, it regulates the high-frequency components obtained through discrete wavelet transform during the sampling process, leading to a different sampling trajectory. Moreover, as the high-frequency components of images significantly affect their visual quality, we can dynamically adjust the weights assigned to the high-frequency guidance loss to balance the reconstruction and perceptual quality. Experimental results demonstrate that our proposed guidance loss significantly enhances performance in the tasks of blind image deblurring, imaging through turbulence, and multiple degradations. Notably, our method achieves a remarkable improvement in PSNR scores by 3.72\,dB. Compared to other state-of-the-art methods, our approach shows a superior capability in generating images with rich details and reduced distortion.
\clearpage  

\section*{Acknowledgements}
This work was supported by the Hong Kong Research Grants Council (RGC) Research Impact Fund (RIF) under Grant R5001-18.
%
%
\bibliographystyle{splncs04}
\bibliography{main}

\begin{thebibliography}{10}
\providecommand{\url}[1]{\texttt{#1}}
\providecommand{\urlprefix}{URL }
\providecommand{\doi}[1]{https://doi.org/#1}

\bibitem{chen2019real}
Chen, C., Xiong, Z., Tian, X., Zha, Z.J., Wu, F.: Real-world image denoising with deep boosting. IEEE Transactions on Pattern Analysis and Machine Intelligence  \textbf{42}(12),  3071--3087 (2019)

\bibitem{cheng2023null}
Cheng, X., Zhang, N., Yu, J., Wang, Y., Li, G., Zhang, J.: Null-space diffusion sampling for zero-shot point cloud completion. In: Proceedings of the Thirty-Second International Joint Conference on Artificial Intelligence (IJCAI). vol.~2 (2023)

\bibitem{cho2021rethinking}
Cho, S.J., Ji, S.W., Hong, J.P., Jung, S.W., Ko, S.J.: Rethinking coarse-to-fine approach in single image deblurring. In: Proceedings of the IEEE/CVF international conference on computer vision. pp. 4641--4650 (2021)

\bibitem{choi2021ilvr}
Choi, J., Kim, S., Jeong, Y., Gwon, Y., Yoon, S.: Ilvr: Conditioning method for denoising diffusion probabilistic models. arXiv preprint arXiv:2108.02938  (2021)

\bibitem{chung2023parallel}
Chung, H., Kim, J., Kim, S., Ye, J.C.: Parallel diffusion models of operator and image for blind inverse problems. In: Proceedings of the IEEE/CVF Conference on Computer Vision and Pattern Recognition. pp. 6059--6069 (2023)

\bibitem{chung2022diffusion}
Chung, H., Kim, J., Mccann, M.T., Klasky, M.L., Ye, J.C.: Diffusion posterior sampling for general noisy inverse problems. ICLR  (2022)

\bibitem{chung2022improving}
Chung, H., Sim, B., Ryu, D., Ye, J.C.: Improving diffusion models for inverse problems using manifold constraints. Advances in Neural Information Processing Systems  \textbf{35},  25683--25696 (2022)

\bibitem{chung2022come}
Chung, H., Sim, B., Ye, J.C.: Come-closer-diffuse-faster: Accelerating conditional diffusion models for inverse problems through stochastic contraction. In: Proceedings of the IEEE/CVF Conference on Computer Vision and Pattern Recognition. pp. 12413--12422 (2022)

\bibitem{cui2023image}
Cui, Y., Ren, W., Cao, X., Knoll, A.: Image restoration via frequency selection. IEEE Transactions on Pattern Analysis and Machine Intelligence  (2023)

\bibitem{deng2019wavelet}
Deng, X., Yang, R., Xu, M., Dragotti, P.L.: Wavelet domain style transfer for an effective perception-distortion tradeoff in single image super-resolution. In: Proceedings of the IEEE/CVF international conference on computer vision. pp. 3076--3085 (2019)

\bibitem{dhariwal2021diffusion}
Dhariwal, P., Nichol, A.: Diffusion models beat gans on image synthesis. Advances in neural information processing systems  \textbf{34},  8780--8794 (2021)

\bibitem{fabian2023diracdiffusion}
Fabian, Z., Tinaz, B., Soltanolkotabi, M.: Diracdiffusion: Denoising and incremental reconstruction with assured data-consistency. arXiv preprint arXiv:2303.14353  (2023)

\bibitem{fei2023generative}
Fei, B., Lyu, Z., Pan, L., Zhang, J., Yang, W., Luo, T., Zhang, B., Dai, B.: Generative diffusion prior for unified image restoration and enhancement. In: Proceedings of the IEEE/CVF Conference on Computer Vision and Pattern Recognition. pp. 9935--9946 (2023)

\bibitem{feng2023score}
Feng, B.T., Smith, J., Rubinstein, M., Chang, H., Bouman, K.L., Freeman, W.T.: Score-based diffusion models as principled priors for inverse imaging. In: Proceedings of the IEEE/CVF International Conference on Computer Vision. pp. 10520--10531 (2023)

\bibitem{ho2020denoising}
Ho, J., Jain, A., Abbeel, P.: Denoising diffusion probabilistic models. Advances in neural information processing systems  \textbf{33},  6840--6851 (2020)

\bibitem{jin2021neutralizing}
Jin, D., Chen, Y., Lu, Y., Chen, J., Wang, P., Liu, Z., Guo, S., Bai, X.: Neutralizing the impact of atmospheric turbulence on complex scene imaging via deep learning. Nature Machine Intelligence  \textbf{3}(10),  876--884 (2021)

\bibitem{kawar2022denoising}
Kawar, B., Elad, M., Ermon, S., Song, J.: Denoising diffusion restoration models. Advances in Neural Information Processing Systems  \textbf{35},  23593--23606 (2022)

\bibitem{kawar2022jpeg}
Kawar, B., Song, J., Ermon, S., Elad, M.: Jpeg artifact correction using denoising diffusion restoration models. arXiv preprint arXiv:2209.11888  (2022)

\bibitem{kawar2021snips}
Kawar, B., Vaksman, G., Elad, M.: Snips: Solving noisy inverse problems stochastically. Advances in Neural Information Processing Systems  \textbf{34},  21757--21769 (2021)

\bibitem{kupyn2019deblurgan}
Kupyn, O., Martyniuk, T., Wu, J., Wang, Z.: Deblurgan-v2: Deblurring (orders-of-magnitude) faster and better. In: Proceedings of the IEEE/CVF international conference on computer vision. pp. 8878--8887 (2019)

\bibitem{laine2019high}
Laine, S., Karras, T., Lehtinen, J., Aila, T.: High-quality self-supervised deep image denoising. Advances in Neural Information Processing Systems  \textbf{32} (2019)

\bibitem{li2023fsr}
Li, J., Dai, T., Zhu, M., Chen, B., Wang, Z., Xia, S.T.: Fsr: A general frequency-oriented framework to accelerate image super-resolution networks. In: Proceedings of the AAAI Conference on Artificial Intelligence. pp. 1343--1350 (2023)

\bibitem{liu2018multi}
Liu, P., Zhang, H., Zhang, K., Lin, L., Zuo, W.: Multi-level wavelet-cnn for image restoration. In: Proceedings of the IEEE conference on computer vision and pattern recognition workshops. pp. 773--782 (2018)

\bibitem{lugmayr2022repaint}
Lugmayr, A., Danelljan, M., Romero, A., Yu, F., Timofte, R., Van~Gool, L.: Repaint: Inpainting using denoising diffusion probabilistic models. In: Proceedings of the IEEE/CVF conference on computer vision and pattern recognition. pp. 11461--11471 (2022)

\bibitem{magid2021dynamic}
Magid, S.A., Zhang, Y., Wei, D., Jang, W.D., Lin, Z., Fu, Y., Pfister, H.: Dynamic high-pass filtering and multi-spectral attention for image super-resolution. In: Proceedings of the IEEE/CVF International Conference on Computer Vision. pp. 4288--4297 (2021)

\bibitem{pan2016l_0}
Pan, J., Hu, Z., Su, Z., Yang, M.H.: $ l\_0 $-regularized intensity and gradient prior for deblurring text images and beyond. IEEE transactions on pattern analysis and machine intelligence  \textbf{39}(2),  342--355 (2016)

\bibitem{pan2017deblurring}
Pan, J., Sun, D., Pfister, H., Yang, M.H.: Deblurring images via dark channel prior. IEEE transactions on pattern analysis and machine intelligence  \textbf{40}(10),  2315--2328 (2017)

\bibitem{qiao2023learning}
Qiao, Y., Shao, M., Wang, L., Zuo, W.: Learning depth-density priors for fourier-based unpaired image restoration. IEEE Transactions on Circuits and Systems for Video Technology  (2023)

\bibitem{ren2020neural}
Ren, D., Zhang, K., Wang, Q., Hu, Q., Zuo, W.: Neural blind deconvolution using deep priors. In: Proceedings of the IEEE/CVF conference on computer vision and pattern recognition. pp. 3341--3350 (2020)

\bibitem{sohl2015deep}
Sohl-Dickstein, J., Weiss, E., Maheswaranathan, N., Ganguli, S.: Deep unsupervised learning using nonequilibrium thermodynamics. In: International conference on machine learning. pp. 2256--2265. PMLR (2015)

\bibitem{song2022pseudoinverse}
Song, J., Vahdat, A., Mardani, M., Kautz, J.: Pseudoinverse-guided diffusion models for inverse problems. In: International Conference on Learning Representations (2022)

\bibitem{sun2021deep}
Sun, H., Bouman, K.L.: Deep probabilistic imaging: Uncertainty quantification and multi-modal solution characterization for computational imaging. In: Proceedings of the AAAI Conference on Artificial Intelligence. pp. 2628--2637 (2021)

\bibitem{sun2022alpha}
Sun, H., Bouman, K.L., Tiede, P., Wang, J.J., Blunt, S., Mawet, D.: $\alpha$-deep probabilistic inference ($\alpha$-dpi): efficient uncertainty quantification from exoplanet astrometry to black hole feature extraction. The Astrophysical Journal  \textbf{932}(2), ~99 (2022)

\bibitem{tsai2022stripformer}
Tsai, F.J., Peng, Y.T., Lin, Y.Y., Tsai, C.C., Lin, C.W.: Stripformer: Strip transformer for fast image deblurring. In: European Conference on Computer Vision. pp. 146--162. Springer (2022)

\bibitem{wang2022zero}
Wang, Y., Yu, J., Zhang, J.: Zero-shot image restoration using denoising diffusion null-space model. arXiv preprint arXiv:2212.00490  (2022)

\bibitem{wang2020deep}
Wang, Z., Chen, J., Hoi, S.C.: Deep learning for image super-resolution: A survey. IEEE transactions on pattern analysis and machine intelligence  \textbf{43}(10),  3365--3387 (2020)

\bibitem{welling2011bayesian}
Welling, M., Teh, Y.W.: Bayesian learning via stochastic gradient langevin dynamics. In: Proceedings of the 28th international conference on machine learning (ICML-11). pp. 681--688. Citeseer (2011)

\bibitem{wu2022blind}
Wu, S., Dong, C., Qiao, Y.: Blind image restoration based on cycle-consistent network. IEEE Transactions on Multimedia  \textbf{25},  1111--1124 (2022)

\bibitem{xiao2023online}
Xiao, J., Jiang, X., Zheng, N., Yang, H., Yang, Y., Yang, Y., Li, D., Lam, K.M.: Online video super-resolution with convolutional kernel bypass grafts. IEEE Transactions on Multimedia  (2023)

\bibitem{xiao2021balanced}
Xiao, J., Liu, T., Zhao, R., Lam, K.M.: Balanced distortion and perception in single-image super-resolution based on optimal transport in wavelet domain. Neurocomputing  \textbf{464},  408--420 (2021)

\bibitem{xiao2019deep}
Xiao, J., Zhao, R., Lai, S.C., Jia, W., Lam, K.M.: Deep progressive convolutional neural network for blind super-resolution with multiple degradations. In: 2019 IEEE International Conference on Image Processing (ICIP). pp. 2856--2860. IEEE (2019)

\bibitem{xiao2021bayesian}
Xiao, J., Zhao, R., Lam, K.M.: Bayesian sparse hierarchical model for image denoising. Signal Processing: Image Communication  \textbf{96},  116299 (2021)

\bibitem{xie2021learning}
Xie, W., Song, D., Xu, C., Xu, C., Zhang, H., Wang, Y.: Learning frequency-aware dynamic network for efficient super-resolution. In: Proceedings of the IEEE/CVF International Conference on Computer Vision. pp. 4308--4317 (2021)

\bibitem{yoo2018image}
Yoo, J., Lee, S.h., Kwak, N.: Image restoration by estimating frequency distribution of local patches. In: Proceedings of the IEEE conference on computer vision and pattern recognition. pp. 6684--6692 (2018)

\bibitem{zamir2021multi}
Zamir, S.W., Arora, A., Khan, S., Hayat, M., Khan, F.S., Yang, M.H., Shao, L.: Multi-stage progressive image restoration. In: Proceedings of the IEEE/CVF conference on computer vision and pattern recognition. pp. 14821--14831 (2021)

\bibitem{zhang2023towards}
Zhang, G., Ji, J., Zhang, Y., Yu, M., Jaakkola, T., Chang, S.: Towards coherent image inpainting using denoising diffusion implicit models (2023)

\bibitem{zhang2020deblurring}
Zhang, K., Luo, W., Zhong, Y., Ma, L., Stenger, B., Liu, W., Li, H.: Deblurring by realistic blurring. In: Proceedings of the IEEE/CVF conference on computer vision and pattern recognition. pp. 2737--2746 (2020)

\bibitem{zhang2018unreasonable}
Zhang, R., Isola, P., Efros, A.A., Shechtman, E., Wang, O.: The unreasonable effectiveness of deep features as a perceptual metric. In: Proceedings of the IEEE conference on computer vision and pattern recognition. pp. 586--595 (2018)

\end{thebibliography}
\end{document}